\documentclass[letterpaper, 10 pt, conference]{ieeeconf}  

\IEEEoverridecommandlockouts                              

\usepackage{graphics} 
\usepackage{epsfig} 
\usepackage{mathptmx} 
\usepackage{times} 
\usepackage{amsmath} 

\usepackage{gensymb}

\usepackage{cite}
\usepackage{textcomp}
\usepackage{amsmath,amssymb,amsfonts}
\usepackage{multicol}
\usepackage{float} 
\usepackage{tablefootnote}
\usepackage{url}
\usepackage{hyperref}
\usepackage{subfigure}
\usepackage{multirow}
\usepackage{array}
\usepackage{xcolor}
\newcolumntype{P}[1]{>{\centering\arraybackslash}p{#1}}

\usepackage{scalerel}
\usepackage{tikz}
\usetikzlibrary{svg.path}

\definecolor{orcidlogocol}{HTML}{A6CE39}
\tikzset{
  orcidlogo/.pic={
    \fill[orcidlogocol] svg{M256,128c0,70.7-57.3,128-128,128C57.3,256,0,198.7,0,128C0,57.3,57.3,0,128,0C198.7,0,256,57.3,256,128z};
    \fill[white] svg{M86.3,186.2H70.9V79.1h15.4v48.4V186.2z}
                 svg{M108.9,79.1h41.6c39.6,0,57,28.3,57,53.6c0,27.5-21.5,53.6-56.8,53.6h-41.8V79.1z M124.3,172.4h24.5c34.9,0,42.9-26.5,42.9-39.7c0-21.5-13.7-39.7-43.7-39.7h-23.7V172.4z}
                 svg{M88.7,56.8c0,5.5-4.5,10.1-10.1,10.1c-5.6,0-10.1-4.6-10.1-10.1c0-5.6,4.5-10.1,10.1-10.1C84.2,46.7,88.7,51.3,88.7,56.8z};
  }
}

\newcommand\orcidicon[1]{\href{https://orcid.org/#1}{\mbox{\scalerel*{
\begin{tikzpicture}[yscale=-1,transform shape]
\pic{orcidlogo};
\end{tikzpicture}
}{|}}}}

\usepackage{graphicx} 

\normalmarginpar 
\newcommand{\IEEEpreprintnotice}{
    \rotatebox{90}{
    \footnotesize
    © 2026 IEEE. This work has been submitted to the IEEE for possible publication.
    Copyright may be transferred without notice, after which this version may no longer be accessible.
    }
}

\usepackage{booktabs}

\title{\LARGE \bf
Towards Generative Predictive Display for Vision-Based Teleoperation: A Zero-Shot Benchmark of Off-the-Shelf Video Models
}

\author{Aws Khalil$^{1}$\orcidicon{0000-0001-9139-3900} and Jaerock Kwon$^{2}$\orcidicon{0000-0002-5687-6998}
\thanks{Both authors are with the Department of Electrical and Computer Engineering, University of Michigan - Dearborn, Dearborn, MI, USA.
{\tt\small awskh@umich.edu$^{1}$},
{\tt\small jrkwon@umich.edu$^{2}$}}%
}

\begin{document}
    \maketitle
    \marginpar{\IEEEpreprintnotice} 
    \begin{abstract}

Teleoperation systems are fundamentally limited by communication latency, which degrades situational awareness and control performance. Predictive display aims to mitigate this limitation by presenting an estimate of the current visual state rather than delayed observations. While recent advances in generative video models enable high-quality video synthesis, their suitability for latency-sensitive predictive display remains unclear.
This paper presents a zero-shot benchmark of off-the-shelf generative video models for short-horizon predictive display, without task-specific fine-tuning. We formulate the problem as rollout-based future frame prediction and develop a unified benchmarking pipeline using simulated driving data from the CARLA simulator. Five publicly released video models spanning transformer-based and diffusion-based families are evaluated across two resolutions and two conditioning regimes (multi-frame and single-frame).
Performance is assessed using prediction accuracy (mean absolute difference), per-rollout latency, peak GPU memory usage, and temporal error evolution across the prediction horizon. On this zero-shot benchmark, no tested model simultaneously achieves low rollout error, non-divergent per-step error behavior, and real-time inference at the source frame rate. Increasing model scale or resolution yields limited and, in some cases, inverted improvements.
These findings highlight a gap between general-purpose generative video synthesis and the requirements of predictive display in teleoperation, suggesting that practical deployment will require either explicit short-horizon temporal supervision, in-domain adaptation, or aggressive inference optimization rather than direct application of off-the-shelf models. Code, configurations, and qualitative results are released on the project page: \url{https://bimilab.github.io/paper-GenPD}

\end{abstract}
    
    \section{Introduction}
\label{sec:introduction}

Teleoperation enables human operators to remotely control robotic and autonomous systems in applications such as autonomous driving, remote inspection, and operation in hazardous environments. However, its performance is fundamentally constrained by communication latency, which causes the operator to perceive outdated visual observations of the environment. Even moderate latency can degrade situational awareness, reduce control accuracy, and negatively impact safety.

A common approach to mitigate this limitation is \emph{predictive display}, where additional information is presented to compensate for delayed observations. In practice, predictive display is often implemented through visual augmentations such as trajectory overlays, free-space visualization, or motion cues. While these techniques improve situational awareness, they do not directly recover the missing visual state of the environment.

An alternative formulation is to estimate the \emph{future visual state} itself. Instead of augmenting delayed observations, the system presents a predicted image intended to approximate the current scene. If sufficiently accurate, such predictions could provide a more direct representation of the environment under latency. In this work, we adopt this formulation and cast predictive display as a short-horizon future frame prediction problem.

Recent advances in generative video modeling have made this direction increasingly relevant. Modern diffusion- and transformer-based models can generate temporally coherent sequences conditioned on visual inputs. However, these models are primarily designed for perceptual realism and long-horizon synthesis, rather than the requirements of latency-sensitive teleoperation.

In predictive display, models must satisfy stricter criteria: predictions must be temporally accurate, consistent across short horizons, and generated under real-time computational constraints. It remains unclear whether current off-the-shelf generative video models meet these requirements.

In this paper, we investigate this question through an empirical zero-shot evaluation of off-the-shelf generative video models for short-horizon predictive display. We formulate the problem as a rollout-based future frame prediction task and evaluate multiple publicly released model families under a unified benchmark protocol. Our analysis considers prediction accuracy, runtime, memory usage, and the temporal evolution of error across the rollout horizon.

\subsection*{Contributions}

\begin{itemize}
    \item We formulate predictive display as a short-horizon rollout-based visual prediction problem under latency constraints, and instantiate it as a unified zero-shot benchmark on simulated driving data.

    \item We benchmark five publicly released generative video models across two resolutions and two conditioning regimes (multi-frame and single-frame), under matched hardware and a controlled inference protocol with no task-specific fine-tuning.

    \item We provide an empirical analysis showing that, in this zero-shot setting and on this benchmark, no tested model simultaneously achieves low rollout MAD, non-divergent per-step rollout error behavior, and real-time inference at the source frame rate. Larger model scale and higher resolution yield mixed and sometimes inverted effects.

    \item We report per-step temporal error in addition to aggregate error and show concrete cases where the two rankings diverge (e.g., SVD's low-resolution drift toward a generic scene), illustrating why aggregate accuracy alone is insufficient for latency-sensitive predictive display.
\end{itemize}

\noindent\textbf{Scope.} This paper deliberately restricts itself to the zero-shot, off-the-shelf setting on a CARLA-based driving benchmark. We do not fine-tune, distill, quantize, or otherwise optimize the deployment of any evaluated model, and we do not study closed-loop or human-in-the-loop teleoperation outcomes. These constitute important follow-up directions and are discussed explicitly in Section~\ref{sec:discussion}.
    \section{Problem Formulation}
\label{sec:problem}

We consider a vision-based teleoperation system in which visual observations from a remote platform are transmitted to a human operator over a communication network. Due to network latency, the operator perceives delayed observations rather than the current state of the environment.

Let $I_t \in \mathbb{R}^{H \times W \times C}$ denote the image captured at time $t$, where $H$, $W$, and $C$ represent image height, width, and number of channels, respectively. Let $\tau$ denote the communication latency. The operator therefore receives $I_{t-\tau}$ instead of the current frame $I_t$.

\subsection{Predictive Display Objective}

The goal of predictive display is to compensate for this latency by estimating the current visual state using past observations. Given a sequence of recent frames, the predictive model aims to generate future frames that approximate the true evolution of the scene.

Let $\Delta$ denote the latency expressed in discrete timesteps (frames). The most recent observation available to the operator is $I_{t-\Delta}$. The predictive model must therefore forecast $\Delta$ steps into the future.

Rather than predicting a single frame, we consider a \emph{rollout formulation} in which the model generates a sequence of future frames:
\begin{equation}
\{\hat{I}_{t-\Delta+1}, \ldots, \hat{I}_{t-\Delta+K}\} = f\left(I_{t-\Delta-k+1}, \ldots, I_{t-\Delta}\right),
\end{equation}
where $K$ denotes the prediction horizon and $f(\cdot)$ is a generative video model.

This formulation enables the system to select the appropriate predicted frame corresponding to the current time, depending on the observed latency.

\subsection{Latency-Aware Frame Selection}

Given a predicted rollout, the frame presented to the operator is selected based on the latency $\Delta$. Concretely, of the $K$ predicted frames produced from the rollout anchored at $I_{t-\Delta}$, the system displays the frame indexed by $\Delta$ steps ahead of that anchor, i.e.\ $\hat{I}_{t-\Delta+\Delta}$, which corresponds to the predicted frame aligned with the current time:
\begin{equation}
\tilde{I}_t \;\triangleq\; \hat{I}_{(t-\Delta)+\Delta},
\qquad 1 \leq \Delta \leq K.
\end{equation}
This decouples prediction from latency estimation: a single rollout supports any observed latency in the range $[1, K]$ frames without retraining the model.

\subsection{Short-Horizon Prediction Setting}

We focus on short-horizon prediction, where $\Delta$ corresponds to tens to hundreds of milliseconds (i.e., a small number of frames). This regime is particularly relevant for teleoperation, where even small temporal errors can significantly impact operator perception and control performance.

Unlike long-horizon video generation, the objective is not to produce visually diverse futures, but to maintain high temporal fidelity relative to the true scene evolution.

\subsection{Evaluation Metrics}

To quantify predictive performance, we evaluate both per-step and aggregate prediction error over the rollout horizon. For each predicted frame $\hat{I}_{t+k}$, the Mean Absolute Difference (MAD) is defined as:
\begin{equation}
\mathrm{MAD}_k = \frac{1}{HWC} \sum_{i=1}^{H} \sum_{j=1}^{W} \sum_{c=1}^{C} 
\left| \hat{I}_{t+k}(i,j,c) - I_{t+k}(i,j,c) \right|.
\end{equation}

We additionally compute the aggregate rollout error as the average over all predicted frames:
\begin{equation}
\mathrm{MAD}_{\text{rollout}} = \frac{1}{K} \sum_{k=1}^{K} \mathrm{MAD}_k.
\end{equation}

This formulation enables analysis of both overall prediction accuracy and the temporal evolution of error.

\subsection{System-Level Constraints}

In addition to prediction accuracy, predictive display systems must satisfy strict computational constraints. Let $T_{\text{inf}}$ denote the average inference time per predicted frame, and $M_{\text{GPU}}$ denote peak GPU memory usage.

For real-time deployment, inference must be sufficiently fast relative to the frame interval. Two regimes are relevant in practice:
\begin{equation}
T_{\text{inf}} < T_{\text{frame}}
\qquad \text{(per-frame streaming)},
\end{equation}
\begin{equation}
T_{\text{roll}} \;\triangleq\; K\,T_{\text{inf}} < K\,T_{\text{frame}}
\qquad \text{(rollout-and-replay)},
\end{equation}
where $T_{\text{frame}}$ is the duration between consecutive frames and $T_{\text{roll}}$ is the end-to-end rollout latency. Generative video models that emit the full $K$-frame rollout in a single inference pass are most naturally evaluated in the rollout-and-replay regime, while autoregressive next-frame predictors are most naturally evaluated in the per-frame streaming regime.

Therefore, the feasibility of predictive display depends on jointly optimizing:
\begin{itemize}
    \item Prediction accuracy (MAD)
    \item Temporal stability of predictions across the rollout
    \item Per-frame inference time $T_{\text{inf}}$ and per-rollout latency $T_{\text{roll}}$
    \item Peak memory usage ($M_{\text{GPU}}$)
\end{itemize}

This work evaluates generative video models under these combined criteria to assess their suitability for latency-sensitive teleoperation systems.
    \section{Related Work}
\label{sec:related-work}

Latency is an inherent challenge in teleoperation systems, arising from delays in sensing, communication, processing, and actuation. In vision-based teleoperation, this latency leads to a temporal mismatch between the current state of the environment and the visual feedback available to the operator. Prior work has addressed this problem through both classical and data-driven approaches, with predictive display emerging as a key technique for improving operator situational awareness.

\subsection{Predictive Display in Teleoperation}

Predictive display has been widely studied as a method for mitigating perception latency by augmenting or reconstructing delayed visual feedback. In early teleoperation systems, predictive displays were primarily based on kinematic prediction and geometric transformations, where future views were rendered using known system motion and camera models \cite{zheng2016experimental, graf2020improving}. These approaches were effective in structured environments such as telemanipulation and remote inspection, but their accuracy was limited by modeling errors and unmodeled dynamics.

In vehicle teleoperation, predictive display has typically been implemented through visual augmentation techniques rather than full scene reconstruction. These include trajectory projection, free-space visualization, and motion extrapolation, which provide the operator with additional cues about the environment \cite{zheng2020evaluation, mirinejad2018modeling, prexl2019motion, hatori2021teleoperation, sato2021implementation, graf2020improving, luz2023enhanced, saparia2021active, prakash2019teleoperated, brudnak2016predictive, prakash2023predictive}. While such methods improve situational awareness, they do not directly recover the missing visual information caused by latency.

More recent work has explored learning-based predictive displays that generate future visual observations. Moniruzzaman \emph{et al.} \cite{moniruzzaman2023long} proposed a predictive display system based on conditional generative adversarial networks (Pix2Pix) combined with optical flow to synthesize future frames. In robotic manipulation environments, generative and perception models have been used for motion prediction and scene understanding \cite{gonzalez2021deserts, sachdeva2021using, xie2021generative}. While these approaches move toward reconstructing future visual states, they are typically developed for structured environments and do not evaluate temporal prediction accuracy under latency constraints in dynamic settings such as vehicle teleoperation.




\subsection{Learning-Based Approaches for Latency Mitigation}

Recent work has explored data-driven approaches for latency mitigation, where models learn delay-compensating dynamics directly from data. Recurrent neural networks (RNNs) \cite{mikolov2010recurrent, naseer2023novel} and Long Short-Term Memory (LSTM) networks \cite{hochreiter1997long, zhang2023predicted, duan2023latency} have been used to capture temporal dependencies and forecast control actions or system trajectories under communication delay \cite{farajiparvar2020brief, boabang2021machine, naseer2023novel}.

In telerobotics, these approaches are often combined with estimation or reinforcement learning techniques. For example, Zhou \emph{et al.} \cite{zhou2022lstm} proposed a bilateral active estimation model (BAEM) using LSTMs for delay compensation, while Bacha \emph{et al.} \cite{bacha2022deep} combined deep deterministic policy gradients (DDPG) with Kalman filtering to improve stability and force feedback. Hybrid approaches have also been explored by integrating reinforcement learning with Model-Mediated Teleoperation (MMT) \cite{beik2020model}.

In vehicle teleoperation, learning-based latency mitigation remains limited. Zhang \emph{et al.} \cite{zhang2023predicted} proposed a Predicted Trajectory Guidance Control (PTGC) system using LSTMs to forecast future trajectories, while Kashwani \emph{et al.} \cite{kashwani2023collision, kashwani2024evaluation} applied transformer-based perception models for object detection and lane segmentation to support predictive overlays.

\subsection{Limitations and Gap}

Across existing approaches, predictive display methods typically fall into two categories: (1) visual augmentation techniques that provide additional cues without reconstructing the full scene, and (2) task-specific learning-based methods that generate future frames under constrained settings.

Importantly, there is no published empirical evaluation of modern, off-the-shelf generative video models for predictive display under a unified rollout protocol. Prior work does not address:
\begin{itemize}
    \item temporal prediction accuracy across a rollout horizon,
    \item the trade-off between predictive fidelity and real-time performance,
    \item model behavior under realistic latency constraints in dynamic environments.
\end{itemize}

This work addresses these gaps by evaluating multiple generative video model families under a unified rollout-based predictive display formulation with explicit analysis of temporal error dynamics and system-level constraints.
    \section{Method}
\label{sec:method}

This section describes the dataset, clip construction, evaluated models, and evaluation protocol used to assess the feasibility of off-the-shelf generative video models for predictive display in teleoperation.

\subsection{Dataset}

We use simulated driving data generated using the CARLA simulator, which provides photorealistic urban environments and realistic vehicle dynamics. The dataset was collected using the MILE framework~\cite{hu2022model}, which utilizes the Roach agent~\cite{zhang2021end} to generate human-like driving behavior. 
Data was collected in Town01 at 15 FPS, reflecting slow urban driving conditions, and consists of front-facing RGB camera streams from an ego vehicle navigating structured road scenarios with dynamic objects, including interactions with other vehicles.

Simulation enables controlled data collection and reproducibility, while preserving key characteristics of real-world driving such as scene geometry, ego-motion, and dynamic object behavior.

\subsection{Clip Construction}

Continuous driving sequences are segmented into short fixed-length clips for evaluation. Each clip consists of:
\begin{itemize}
    \item $9$ conditioning frames $\{I_{t-8}, \ldots, I_t\}$
    \item $8$ future frames $\{I_{t+1}, \ldots, I_{t+8}\}$
\end{itemize}

The conditioning frames represent the most recent observations available to the model, while the future frames define the prediction horizon. This formulation enables evaluation of short-horizon predictive performance under consistent temporal structure.

Clips are sampled from multiple driving sequences to ensure diversity in scene layout, motion patterns, and traffic conditions.

\subsection{Resolutions}

To analyze the impact of spatial resolution on predictive accuracy and computational cost, experiments are conducted at two resolutions:
\begin{itemize}
    \item $256 \times 160$
    \item $512 \times 320$
\end{itemize}

These resolutions are selected to study the trade-off between computational efficiency and spatial detail within the constraints of the evaluated models.

\subsection{Evaluated Models}

We evaluate multiple off-the-shelf generative video model families, each with different architectures and conditioning mechanisms:

\textbf{LTX-Video \cite{hacohen2024ltx}:}
\begin{itemize}
    \item LTX 2B (distilled)
    \item LTX 13B
\end{itemize}

\textbf{Stable Video Diffusion (SVD) \cite{blattmann2023stable}:}
\begin{itemize}
    \item SVD 1.1 (image-to-video)
\end{itemize}

\textbf{Wan \cite{wan2025wan}:}
\begin{itemize}
    \item Wan VACE 1.3B (masked multi-frame conditioning)
    \item Wan I2V 1.3B (single-frame image-to-video baseline; Diffusers conversion)
\end{itemize}

All models are used without task-specific fine-tuning, reflecting a realistic deployment scenario where pretrained generative models are directly applied to teleoperation data.

\noindent\textbf{Native vs.\ benchmark resolution.} The two benchmark resolutions ($256{\times}160$ and $512{\times}320$, $16{:}10$ aspect ratio) differ from the native release-time operating resolutions and aspect ratios of all five models, which were primarily trained on higher-resolution, more square-leaning data (typically $\geq 480$p, with model-specific aspect ratios documented in each model card). The benchmark therefore reports zero-shot behavior at \emph{non-native} resolutions; this is a deliberate choice to expose models to the lower-bandwidth regime relevant to teleoperation, but it implies that resolution-trend results, in particular for Wan~VACE, may partially reflect resolution mismatch rather than predictive ability per se. We flag this as a confound rather than as a controlled variable.

\subsection{Inference Setup}

For each clip, a sequence of $9$ conditioning frames is extracted. Depending on the model architecture, these frames are provided either fully or partially as input. Specifically, multi-frame-conditioned models (LTX-2B, LTX-13B, and Wan VACE) utilize the full sequence, while image-to-video models (SVD and Wan I2V) condition only on the final frame of the sequence.

Each model generates a sequence of $8$ predicted frames:
\[
\{\hat{I}_{t+1}, \ldots, \hat{I}_{t+8}\}.
\]

To ensure a consistent evaluation interface across models with different conditioning mechanisms, all outputs are formatted into a unified sequence consisting of:
\begin{itemize}
    \item the original $9$ conditioning frames
    \item followed by $8$ generated rollout frames
\end{itemize}

\noindent\textbf{Conditioning regime as a confound.} Because the multi-frame and single-frame regimes differ in the information passed to the model, comparisons across model families combine effects of both \emph{architecture} and \emph{conditioning regime}. We therefore interpret all cross-family results as architecture-plus-conditioning comparisons, and reserve cleaner architecture-only conclusions for within-family pairs (e.g., LTX-2B vs.\ LTX-13B, Wan~I2V vs.\ Wan~VACE).

Inference is performed on a single GPU under consistent hardware conditions. We report the per-frame inference time $T_{\text{inf}}$ as the total wall-clock time of one rollout call divided by the $8$ generated frames; the corresponding end-to-end rollout latency is $T_{\text{roll}} = 8\,T_{\text{inf}}$, which is the relevant quantity for systems that consume an entire rollout before refreshing. Reported timings exclude model load time and use a warm GPU (one untimed warm-up call per model). Peak GPU memory is read once per run via the deterministic CUDA peak-memory API after the warm-up call; consequently the peak-memory column is effectively a single deterministic measurement and is reported without a standard deviation in Section~\ref{sec:results}.

\subsection{Evaluation Metrics}

We evaluate model performance using three complementary metrics:

\subsubsection{Prediction Accuracy}

Prediction accuracy is measured using the Mean Absolute Difference (MAD) between predicted and ground-truth frames across the rollout horizon. Pixels are evaluated in $8$-bit RGB space ($[0,255]$), averaged jointly over height, width, and channel: With $k = 1,\ldots,8$,
\[
\text{MAD}_k = \frac{1}{HWC} \sum_{i=1}^{H}\sum_{j=1}^{W}\sum_{c=1}^{C} \left| \hat{I}_{t+k}(i,j,c) - I_{t+k}(i,j,c) \right|, 
\]
which matches the definition introduced in Eq.~(3) of Section~\ref{sec:problem}. We report both the per-step $\text{MAD}_k$ and the aggregate rollout MAD averaged over the $K\!=\!8$ predicted frames.

We acknowledge that pixel-level MAD is a coarse proxy for predictive utility: two predictions with the same MAD can differ significantly in perceptual fidelity and in task relevance. Within the scope of this zero-shot benchmark, MAD is used as a uniform, model-agnostic prediction-error indicator that is directly comparable across all evaluated architectures; perceptual (SSIM, LPIPS) and task-aware metrics (e.g., lane/road-mask consistency, optical-flow consistency) are deferred to future work and discussed in Section~\ref{sec:discussion}.

\subsubsection{Runtime Performance}

Runtime is measured as $T_{\text{inf}}$ in seconds per generated frame and, equivalently, as the per-rollout latency $T_{\text{roll}} = 8\,T_{\text{inf}}$. Both quantities are reported under the same warm-GPU protocol described above. For real-time deployment at the source frame rate, the relevant constraint is $T_{\text{inf}} < T_{\text{frame}}$ (per-frame streaming) or $T_{\text{roll}} < K \cdot T_{\text{frame}}$ (rollout-and-replay). At the source rate of 15~FPS, $T_{\text{frame}} = 66.7$~ms and $K \cdot T_{\text{frame}} \approx 533$~ms.

\subsubsection{Memory Usage}

Peak GPU memory consumption during inference is recorded to assess deployment feasibility on resource-constrained platforms.

\subsection{Evaluation Protocol}

All models are evaluated across:
\begin{itemize}
    \item multiple clips sampled from different driving sequences
    \item two spatial resolutions
\end{itemize}

For each configuration, we report aggregate metrics as well as temporal error evolution across the prediction horizon. This allows us to analyze not only prediction accuracy, but also the stability and consistency of predictions over time.

Unlike single-frame evaluation, this rollout-based protocol captures temporal error dynamics, which are critical for predictive display applications where the operator relies on consistent visual feedback under latency.
    \section{Experimental Setup}
\label{sec:experimental-setup}

All experiments were conducted on a workstation running Ubuntu 22.04, equipped with an AMD Ryzen Threadripper Pro 5975WX (32 cores / 64 threads) CPU and an NVIDIA RTX 6000 Ada GPU with 48\,GB of VRAM.

All models were evaluated under consistent hardware conditions to ensure fair comparison of runtime and memory usage. No model-specific hardware optimizations or distributed inference strategies were employed. Each experiment was executed using a single GPU. Inference was performed using PyTorch and Hugging Face Diffusers frameworks, depending on the model implementation.

\noindent\textbf{Sampling and seeds.} For each model we use the publicly released checkpoint and the inference defaults shipped with its reference implementation (sampler, number of denoising steps, classifier-free-guidance scale where applicable). For diffusion-based models, a fixed random seed is set at the start of each rollout for reproducibility; for transformer-based models with deterministic decoding the result is deterministic given the conditioning frames.

\noindent\textbf{Runtime measurement.} Per-frame inference time $T_{\text{inf}}$ is computed as the wall-clock time of a single rollout call divided by the $K\!=\!8$ generated frames, after one untimed warm-up call per model to amortise CUDA context creation and weight loading. We report mean and standard deviation of $T_{\text{inf}}$ across the evaluated rollouts; the corresponding end-to-end rollout latency is $T_{\text{roll}} = 8\,T_{\text{inf}}$.

\noindent\textbf{Memory measurement.} Peak GPU memory is read once per run via the CUDA peak-memory API (\texttt{torch.cuda.max\_memory\_allocated}) after the warm-up call. Because the value is deterministic for a given model, resolution, and sampler configuration, it is reported as a single peak number without a standard deviation in Table~\ref{tab:main_results}. This is the operationally meaningful quantity for deployment feasibility.
    \section{Results}
\label{sec:results}

We evaluate the feasibility of off-the-shelf generative video models for predictive display in teleoperation under a standardized rollout protocol. Each model is conditioned on $9$ observed frames and tasked with generating $8$ future frames. Performance is assessed using Mean Absolute Difference (MAD), runtime (seconds per frame), and GPU memory usage. In addition to aggregate metrics, we analyze the temporal evolution of prediction error across the rollout horizon.

\subsection{Quantitative Performance}

Table~\ref{tab:main_results} summarizes the aggregate performance across all models and resolutions. Among all evaluated models, the LTX family consistently achieves the lowest prediction error, with both LTX-2B and LTX-13B substantially outperforming SVD and the Wan-based models on aggregate MAD.

Within the LTX family, the smaller distilled LTX-2B in fact achieves \emph{lower} aggregate MAD than the larger LTX-13B at both resolutions ($14.23$ vs.\ $22.34$ at $256{\times}160$; $11.55$ vs.\ $19.57$ at $512{\times}320$). This is a reversal rather than a smooth diminishing-returns curve: under the off-the-shelf zero-shot protocol used here, scaling from $2$B to $13$B does not buy a better short-horizon predictor on this benchmark. We report this as an empirical observation. A plausible (but unverified) hypothesis is that LTX-2B's distillation produces fewer denoising steps and shorter, more deterministic generations that incidentally align well with rollout-style short-horizon prediction, whereas LTX-13B targets longer-horizon, higher-fidelity synthesis. We do not claim to explain the reversal here; isolating the contributions of distillation, sampling-step budget, and native operating resolution would require a controlled ablation and is left to future work.

In contrast, SVD exhibits the highest error despite achieving the fastest runtime, indicating that lower latency alone does not guarantee usable predictive quality. Wan I2V provides intermediate performance, achieving moderate error with lower runtime than LTX, while Wan VACE performs poorly at lower resolution and improves substantially at higher resolution. We caution that part of this Wan VACE trend may be attributable to a mismatch between the benchmark resolutions and each model's native training resolution, rather than to predictive ability per se; aggregate runtime and memory therefore also shift with resolution in the Wan VACE row.

\begin{table*}
\begin{tabular*}{\textwidth}{@{\extracolsep{\fill}} llcccc}
\toprule
Model & Resolution & MAD & Sec/Frame & Rollout (s) & VRAM (GB) \\
\midrule
LTX 13B & 256x160 & 22.34 $\pm$ 3.32 & 3.73 $\pm$ 0.10 & 29.84 & 46.68 \\
LTX 13B & 512x320 & 19.57 $\pm$ 2.10 & 3.85 $\pm$ 0.08 & 30.80 & 46.68 \\
LTX 2B & 256x160 & 14.23 $\pm$ 3.83 & 3.12 $\pm$ 0.07 & 24.96 & 25.97 \\
LTX 2B & 512x320 & 11.55 $\pm$ 2.67 & 3.17 $\pm$ 0.09 & 25.36 & 25.97 \\
SVD 1.1 & 256x160 & 78.68 $\pm$ 11.00 & 1.31 $\pm$ 0.06 & 10.48 & 3.15 \\
SVD 1.1 & 512x320 & 83.24 $\pm$ 26.10 & 1.54 $\pm$ 0.05 & 12.32 & 3.44 \\
Wan I2V 1.3B & 256x160 & 26.90 $\pm$ 2.81 & 2.65 $\pm$ 0.07 & 21.20 & 10.80 \\
Wan I2V 1.3B & 512x320 & 23.52 $\pm$ 6.29 & 2.92 $\pm$ 0.05 & 23.36 & 10.80 \\
Wan VACE 1.3B & 256x160 & 89.42 $\pm$ 4.59 & 2.54 $\pm$ 0.04 & 20.32 & 11.70 \\
Wan VACE 1.3B & 512x320 & 32.08 $\pm$ 6.29 & 3.82 $\pm$ 0.09 & 30.56 & 13.58 \\
\bottomrule
\end{tabular*}

\vspace{0.4em}
\noindent\footnotesize MAD reported as mean $\pm$ standard deviation across evaluated clips. \emph{Sec/Frame} is the per-frame inference time $T_{\text{inf}}$ (mean $\pm$ std across rollouts on a warm GPU); \emph{Rollout (s)} is the corresponding end-to-end rollout latency $T_{\text{roll}} = 8\,T_{\text{inf}}$. \emph{VRAM (GB)} is the deterministic peak GPU memory read once per run via the CUDA peak-memory API and is therefore reported without a standard deviation. The frame budget at the source rate of 15~FPS is $T_{\text{frame}} \approx 66.7$~ms; all reported $T_{\text{inf}}$ values exceed this budget by $20$--$60\times$.
\caption{
Aggregate performance across models, reporting mean absolute difference (MAD), inference time per frame, and peak GPU memory usage across all evaluated clips.
}
\label{tab:main_results}
\end{table*}

\subsection{Quality--Speed Tradeoff}

\begin{figure}
\includegraphics[width=\linewidth]{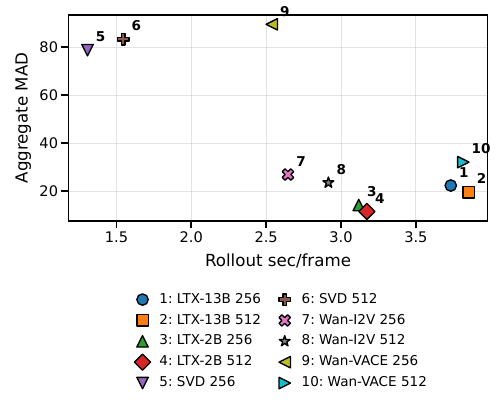}
\caption{Tradeoff between prediction accuracy (MAD) and inference time across models and resolutions. Each point corresponds to a model-resolution configuration.}
\label{fig:tradeoff}
\end{figure}

Figure~\ref{fig:tradeoff} illustrates the tradeoff between prediction accuracy and computational cost. None of the evaluated models simultaneously achieves low error and low runtime. While SVD operates at the lowest latency, its high error renders it unsuitable for predictive display. LTX models provide the best accuracy but incur higher computational cost. Wan I2V occupies a middle ground, offering a compromise between accuracy and runtime, though still falling short of LTX performance.

These results highlight a fundamental limitation of current generative video models: improving predictive fidelity generally requires increased computational resources, which conflicts with real-time constraints in teleoperation.

\subsection{Temporal Error Behavior}

\begin{figure}
\includegraphics[width=\linewidth]{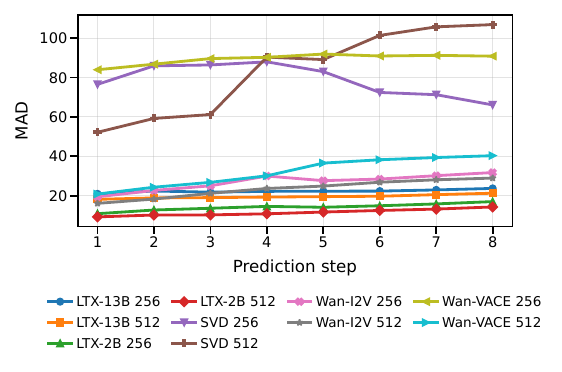}
\caption{Temporal rollout error across prediction steps.}
\label{fig:temporal_error}
\end{figure}

To better understand model behavior over time, we analyze the per-step MAD across the prediction horizon, as shown in Figure~\ref{fig:temporal_error}. Distinct temporal patterns emerge across model families.

LTX models exhibit relatively stable error profiles, with only a mild increase in error as the prediction horizon extends. This suggests that LTX maintains temporal consistency and is capable of short-horizon forecasting.

In contrast, SVD demonstrates inconsistent temporal behavior. At lower resolution, the per-step error \emph{decreases} over the horizon despite an already-large initial error -- a pattern consistent with convergence toward a low-frequency, scene-agnostic representation rather than improving prediction. We note that this interpretation is consistent with the qualitative observation in Figs.~\ref{fig:qual-clip1}--\ref{fig:qual-clip2}, where SVD predictions visibly drift away from the conditioning scene structure. At higher resolution, SVD error instead increases sharply, indicating instability and divergence from the ground truth.

Wan I2V shows a consistent increase in error over time, reflecting gradual drift from the true future frames. While its predictions retain some structural coherence, temporal accuracy degrades steadily. Wan VACE exhibits high initial error and further degradation over time, particularly at higher resolution, indicating difficulty in leveraging multi-frame conditioning for predictive tasks.

\subsection{Effect of Resolution}

Increasing spatial resolution from $256\times160$ to $512\times320$ yields mixed effects across models. For some models, such as Wan VACE, higher resolution improves aggregate error, suggesting better preservation of coarse spatial structure. However, this improvement does not translate to stable temporal predictions, as error still increases over time. For other models, such as SVD, higher resolution exacerbates instability, leading to significantly worse temporal behavior.

Overall, increasing resolution alone is insufficient to address the core limitations of predictive performance.

\subsection{Qualitative Analysis}

\begin{figure*}
\includegraphics[width=\linewidth]{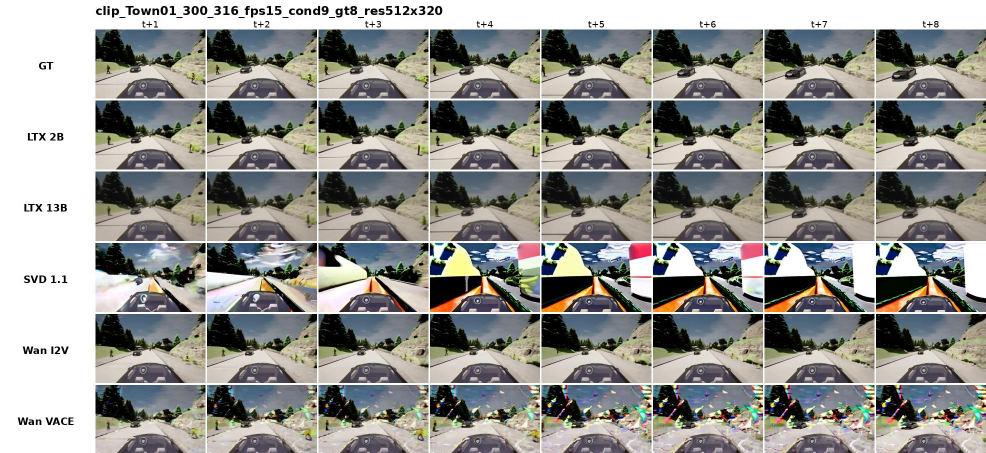}
\caption{Qualitative comparison on a representative driving clip at $512 \times 320$. Each row shows an $8$-step rollout for a different model given the same conditioning frames.}
\label{fig:qual-clip1}
\end{figure*}

\begin{figure*}
\includegraphics[width=\linewidth]{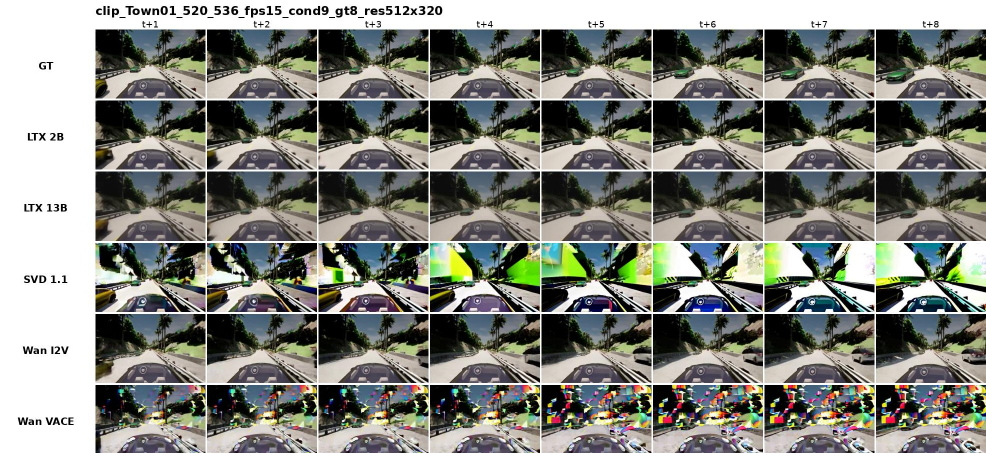}
\caption{Qualitative comparison on a second driving clip with more dynamic scene elements. Predictions are shown at $512 \times 320$ to highlight spatial and temporal inconsistencies across models.}
\label{fig:qual-clip2}
\end{figure*}

To complement the quantitative evaluation, we present qualitative comparisons in Figures~\ref{fig:qual-clip1} and~\ref{fig:qual-clip2}. Each figure shows an $8$-step rollout for multiple models under identical conditioning inputs. The selected clips are representative examples chosen to illustrate typical model behavior, including both relatively structured and more dynamic driving scenarios. All qualitative results are shown at the higher resolution of $512 \times 320$ to better expose spatial artifacts and temporal inconsistencies.

Across both clips, the LTX models produce the most visually consistent predictions, preserving scene layout and object structure over the prediction horizon. While slight blurring is observed, the overall geometry of the road, surrounding environment, and dynamic agents remains stable. In contrast, SVD exhibits severe degradation, with predictions quickly diverging into unrealistic and fragmented patterns, particularly in regions involving motion or structural detail. This behavior aligns with the high error observed in the quantitative results.

Wan I2V demonstrates intermediate behavior, maintaining coarse scene structure but showing gradual temporal drift, particularly in dynamic regions such as moving vehicles and pedestrians. Wan VACE performs the worst qualitatively, with strong artifacts and unstable predictions that fail to preserve scene semantics. Notably, increasing resolution reveals these failure modes more clearly, highlighting inconsistencies that are less apparent at lower resolutions.

These qualitative results reinforce the quantitative findings by illustrating that, while some models can generate visually plausible frames, maintaining temporally accurate and structurally consistent predictions remains a significant challenge for predictive display applications.

The selected clips are representative in terms of scene structure and motion complexity, covering both relatively static environments and scenarios with dynamic agents such as vehicles and pedestrians.

    \section{Discussion}
\label{sec:discussion}

The results highlight a gap between general-purpose generative video modeling and the requirements of predictive display in teleoperation, in the zero-shot regime studied here. While several models generate visually plausible outputs, this visual realism does not, on its own, translate into accurate short-horizon prediction at the source frame rate.

Among the evaluated models, the LTX family achieves the lowest error and the most stable temporal behavior on this benchmark, suggesting stronger alignment with predictive objectives. However, these models remain computationally expensive on a single GPU without optimization. Diffusion-based models such as SVD and Wan exhibit higher error and less stable temporal behavior, including drift and non-monotonic error patterns. The within-family LTX-2B vs.\ LTX-13B reversal further suggests that, off-the-shelf, model scale alone is not a reliable lever for short-horizon predictive quality.

Per-step temporal analysis surfaces failure modes that aggregate MAD does not distinguish. Most concretely, SVD at $256{\times}160$ has an aggregate MAD that is high but \emph{decreasing} over the horizon, which is consistent with convergence to a low-frequency scene-agnostic representation rather than improving prediction. A practitioner inspecting only aggregate MAD or only end-of-horizon MAD would draw a different conclusion than one inspecting the full per-step curve. We report this as a useful empirical observation about how to interpret rollout error, rather than as a new metric contribution.

From a systems perspective, none of the evaluated models, in their off-the-shelf zero-shot configuration on this benchmark, satisfy the combined requirements of accuracy, temporal consistency, and real-time inference at the 15~FPS source rate. We do not claim that this rules out generative video models for predictive display in general; rather, we read it as evidence that practical deployment will likely require either explicit short-horizon temporal supervision (in-domain fine-tuning or distillation), aggressive inference optimization (compilation, quantization, TensorRT, custom samplers), or architectures explicitly designed for rollout-style prediction under compute constraints.

\subsection{Scope and Limitations}
\label{sec:limitations}

We deliberately keep the scope of this paper narrow, and we list the corresponding limitations explicitly so that the conclusions are not overgeneralized.

\begin{itemize}
    \item \textbf{Zero-shot, off-the-shelf only.} All models are run at their published checkpoints, with no fine-tuning, distillation, quantization, or compilation specific to this work. The benchmark therefore answers ``how do these models behave when applied directly,'' not ``what is the best these architectures can do.''
    \item \textbf{Pixel-level error only.} We use Mean Absolute Difference because it is uniform across architectures and does not require auxiliary models. Perceptual metrics (SSIM, LPIPS) and task-aware metrics (lane/road-mask consistency, optical-flow consistency, feature-track displacement) would provide complementary evidence and are deferred to future work.
    \item \textbf{No closed-loop or human-in-the-loop study.} We measure intrinsic prediction quality but not downstream operator behavior such as lane deviation, collision rate, reaction time, or task success under predictive display. A latency-in-the-loop study is the most direct way to translate these intrinsic metrics into teleoperation utility.
    \item \textbf{Simulator-only evaluation.} The benchmark uses CARLA Town01 with the MILE/Roach data-collection stack. Real driving footage and additional CARLA towns / weather / agents are out of scope here; transfer of these conclusions to real video should not be assumed without direct verification.
    \item \textbf{No classical or specialised baselines.} We do not compare to last-frame persistence, optical-flow / ego-motion warping, or lightweight learned next-frame predictors, all of which are natural baselines for short-horizon predictive display. Their inclusion would strengthen any claim about \emph{relative} usefulness of generative video models in this regime.
    \item \textbf{Single hardware, single inference stack.} All timings are obtained on one NVIDIA RTX~6000 Ada with a stock PyTorch / Diffusers stack. Real-time conclusions are specific to this configuration and are not robust to model-specific kernel or runtime optimization.
    \item \textbf{Conditioning regime is bundled with architecture.} Multi-frame and single-frame conditioning are not orthogonalised, so cross-family comparisons reflect both architecture and conditioning regime; we report architecture-only conclusions only within families.
    \item \textbf{Limited model coverage.} We evaluate five publicly released models. Other recent video generators (e.g., longer-context diffusion models, prediction-specialised architectures, world models) are not covered, and the conclusions should be read as ``these five models on this benchmark'' rather than as a statement about ``current video models'' as a class.
    \item \textbf{Aggregate-only uncertainty reporting.} Table~\ref{tab:main_results} reports mean and standard deviation of MAD and $T_{\text{inf}}$ across the evaluated rollouts, but does not report exact clip counts per configuration, per-clip distributions, paired tests between model pairs, or 95\% confidence intervals. Per-clip results were not preserved in a form that enables paired analysis after the fact; obtaining this would require a re-run of the benchmark with structured per-clip logging, which we view as a necessary upgrade for any future archival benchmark version of this work.
\end{itemize}

These limitations directly motivate the follow-up directions outlined above: in-domain adaptation, perceptual and task-aware metrics, classical baselines, optimised inference, real-data evaluation, and a closed-loop teleoperation study.
    \section{Conclusion}
\label{sec:conclusion}

This paper evaluated the feasibility of off-the-shelf generative video models for predictive display in teleoperation using a unified, zero-shot, short-horizon rollout benchmark on simulated driving data. Across five publicly released models, two resolutions, and two conditioning regimes, we assessed prediction accuracy, per-rollout latency, peak GPU memory, and temporal error behavior.

Within the scope defined in Section~\ref{sec:limitations}, our results show that none of the tested off-the-shelf models simultaneously achieve low rollout MAD, non-divergent per-step rollout error behavior, and real-time inference at the source frame rate. Models with relatively lower error remain computationally expensive on a single GPU; faster models exhibit substantially higher error or unstable temporal behavior. Per-step temporal analysis additionally reveals failure modes that the aggregate error does not surface, including SVD's drift toward a generic, scene-agnostic representation at lower resolution.

These findings should be interpreted narrowly: they characterise off-the-shelf zero-shot behavior on this benchmark and on this hardware, not the ultimate capability of generative video models for predictive display. Bridging the gap we observe will likely require some combination of in-domain temporal supervision (fine-tuning, distillation), aggressive inference optimization (compilation, quantization, custom samplers), perceptual and task-aware evaluation, and ultimately a closed-loop or human-in-the-loop teleoperation study to ground intrinsic prediction quality in operator outcomes.

    \bibliographystyle{unsrt}
\bibliography{3_end/references.bib,3_end/references-teleop-av}
\end{document}